\documentclass[conference]{IEEEtran}
\IEEEoverridecommandlockouts

\usepackage{cite}
\usepackage{amsmath,amssymb}
\usepackage{graphicx}
\usepackage{booktabs}
\usepackage{multirow}
\usepackage{array}
\usepackage{url}
\usepackage{balance}
\usepackage{xcolor}
\usepackage[normalem]{ulem}
\definecolor{customcolor}{RGB}{0, 0, 238}
\usepackage[colorlinks=true, linkcolor=black, urlcolor=customcolor, citecolor=black]{hyperref}
\usepackage{todonotes}
\graphicspath{{figures/}}

\title{Recursive LLM Degradation in Biomedical Question Answering:
A Cross-Generation Study}

\author{
\IEEEauthorblockN{Bibek Bhandari$^{*}$}
\IEEEauthorblockA{
Department of Computer Science and Engineering\\
CMR Institute of Technology, Bengaluru, India\\
bibekbhandari9848@gmail.com
}
\and
\IEEEauthorblockN{Kshitij Lingthep$^{*}$}
\IEEEauthorblockA{
Independent Researcher\\
Memphis, Tennessee, USA\\
kshitij.lingthep.cs@outlook.com
}
\thanks{$^*$Both authors contributed equally to this work.}
}
\IEEEaftertitletext{\vspace{-2.0em}}

\usepackage{csquotes}
\MakeOuterQuote{"}
\begin{document}
\raggedbottom
\maketitle
\begin{abstract}
Repeatedly training language models on their own generated data may create a synthetic-data feedback loop in which errors and distributional biases are reintroduced into subsequent training datasets. This paper studies that process in biomedical question answering (QA) using PubMedQA and two Qwen2.5 model sizes, 0.5B and 3B parameters. The study compares a recursive synthetic-data condition, in which generation $G_{k+1}$ is trained on answers produced by $G_k$, against a Human-Control condition that repeatedly uses the original human training data. The study evaluates across four generations from $G_0$--$G_3$ with two random seeds (42 and 123) and a fixed evaluation set of 1,000 expert-labeled samples. The evaluation includes disease and chemical entity F1, context-supported rate, lexical and semantic similarity, answer length, repetition rate, and other evaluation metrics. The Recursive condition for both model sizes and both seeds showed larger declines than the Human-Control condition in disease entity F1, chemical entity F1, context-supported rate, ROUGE-L, and cosine similarity. Under the fixed no-repeat 3-gram decoding constraint, the main observed behavioral change was increased answer length, while the measured 3-gram repetition rate did not increase. The magnitude of the difference-in-change was larger for the 3B model than for the 0.5B model. This difference was particularly apparent in disease F1, context-supported rate, cosine similarity, and answer length. These results show domain-specific changes associated with using recursive synthetic-data training in biomedical QA, but do not establish clinical hallucination rates or universal model collapse.
\end{abstract}


\section{Introduction}
Large language models are increasingly used to generate synthetic data that eventually circulates back into future training datasets. In the future, if synthetic data replaces or dominates human-authored data, future model generations may repeatedly train on generated synthetic data. This can cause training data to drift further from the original data distribution and form a synthetic-data feedback loop. This effect has been described as “model collapse” in earlier works because it demonstrated that recursively generated synthetic data can progressively remove information from the original data distribution \cite{shumailov2024collapse}. However, biomedical QA is more challenging because quality cannot be determined solely by generic surface-level similarity. Biomedical answers may remain fluent and coherent even when they omit, alter, or introduce medically relevant entities. This motivates evaluating recursive synthetic-data training with domain-specific metrics in addition to conventional lexical and semantic metrics. 

This paper studies recursive synthetic-data training in PubMedQA, a biomedical QA benchmark that includes expert-labeled samples and biomedical contexts \cite{jin2019pubmedqa}. The study evaluates two Qwen2.5 model sizes, 0.5B and 3B parameters \cite{qwen2024qwen25}, across four generations. The central comparison is between (i) Recursive synthetic-data training, where each generation is trained on answers generated by the previous generation, and (ii) Human-Control condition, where each generation is trained on the original human training set. The study compares changes by using the same evaluation set across generations and conditions.

The study considers four questions: (1) Does recursive synthetic-data training produce measurable changes across generations? (2) Which dimensions of biomedical answer quality are most sensitive to recursive synthetic-data training? (3) Are the observed changes reproducible across random seeds? (4) Does the magnitude of the difference-in-change between the Recursive and Human-Control condition vary with model size?

The main finding is not a uniform deterioration across all metrics. Recursive synthetic-data training produces a coordinated pattern of decreasing disease and chemical entity F1, decreasing context-supported rate, and decreasing lexical and semantic alignment. However, the answer length increases. These changes are larger than those observed under Human-Control condition for the primary metrics and are generally larger in the tested 3B configuration.

\section{Contributions}
This study makes four contributions. (1) It provides a controlled biomedical QA setup for recursive synthetic-data training across $G_0$--$G_3$. (2) It compares recursive synthetic-data training with a Human-Control condition that repeatedly trains on the original human-authored data, providing a control condition that repeatedly trains on fixed human-authored training data. (3) It evaluates changes across biomedical entity, semantic, lexical, and behavioral metrics. The analysis also compares the difference-in-change between the Recursive and Human-Control condition across random seeds and model sizes. (4) The study provides the code and research materials which include datasets, model inferences, and evaluation results, so that the work can be reproduced and further analyzed.

\section{Related Work}
\subsection{Recursive training and synthetic-data feedback loop}
Recent research on recursive synthetic-data training and model collapse, as well as the growing use of synthetic data to train and fine-tune language models, drives this study. Shumailov \textit{et al.} showed that repeatedly training models on recursively generated synthetic data can gradually erode information from the original data distribution \cite{shumailov2024collapse}. Similar "self-consuming" studies examine performance, data quality, and stability across repeated synthetic-data loops \cite{alemohammad2024mad,fu2024selfconsuming,gillman2024selfcorrecting}. More recent work studies conditions under which recursive synthetic-data training collapses or remains stable. These studies also examine how synthetic-data generation and curation affect outcomes across training generations \cite{kazdan2025collapse,zhu2025synthesize,suresh2025rate,fu2025prevent,wei2025adversarial}. This study applies these questions to biomedical QA and focuses on biomedical entities, context-supported rate, and reference-answer alignment in addition to generic text, lexical and semantic similarity.

\subsection{Iterative synthetic-data enhancement}
Self-generated training data has also been used for instruction synthesis and data enhancement \cite{wang2023selfinstruct,lee2024llm2llm}. These approaches demonstrate the utility of synthetic data while also motivating the study of recursive synthetic-data training as a potential bottleneck when synthetic data are repeatedly fed back into subsequent training generations.

\subsection{Biomedical question answering and medical language models}
Recent medical language-model studies examine biomedical and clinical adaptation, including GatorTron, PMC-LLaMA, BioMistral, and MEDITRON \cite{yang2022gatortron,wu2023pmcllama,labrak2024biomistral,chen2023meditron}.  This study uses a biomedical NER model as an auxiliary evaluation instrument rather than as the generation model. Biomedical QA has also been evaluated using expert-labeled biomedical data, as demonstrated by PubMedQA \cite{jin2019pubmedqa}.

\subsection{Models, efficient adaptation, and domain-aware evaluation}
The study uses Qwen2.5 \cite{qwen2024qwen25} with parameter-efficient quantized fine-tuning using QLoRA \cite{dettmers2023qlora}. PubMedBERT is used as the domain-specific language model underlying the NER model for entity analysis \cite{gu2020pubmedbert}. The NER model uses PubMedBERT to identify the disease and chemical entity categories defined in BC5CDR \cite{li2016bc5cdr}. In addition to these domain-specific models, the study evaluates results using reference-answer alignment and context-support measurements.

\section{Experimental Design}
\subsection{Dataset and fixed evaluation set}
The study uses the PubMedQA dataset to evaluate changes in biomedical QA performance \cite{jin2019pubmedqa}. The training data consists of 5,000 samples from the PQA-U split, while 1,000 samples from the PQA-L split are reserved for evaluation. The same evaluation set with 1000 samples is used across all generations, training conditions, random seeds, and model sizes. Training and evaluation sets are checked for identifier overlap before model training to prevent evaluation data from entering the training pool.

\subsection{Recursive and Human-Control conditions}
Let $D_0$ denote the original human-authored training set and $M_g$ denote the model at generation $g$. In the Recursive condition,
\begin{equation}
D_{g+1}=G(M_g, D_0),
\end{equation}
where $G$ denotes generation of answers for the training questions and contexts. The next model is fine-tuned on $D_{g+1}$ using the previous generation's adapter as the initialization. Thus the Recursive synthetic-data training process is:
\begin{equation}
D_0\rightarrow M_0\rightarrow D_1\rightarrow M_1\rightarrow D_2\rightarrow M_2\rightarrow D_3\rightarrow M_3.
\end{equation}
In the Human-Control condition, each generation is fine-tuned on the original human-authored training set $D_0$. So, the training data is fixed and it removes the recursive synthetic-data training scenario. Thus the Human-Control training process is:
\begin{equation}
D_0\rightarrow M_0\rightarrow D_0\rightarrow M_1\rightarrow D_0\rightarrow M_2\rightarrow D_0\rightarrow M_3.
\end{equation}

\subsection{Models and training}
The study evaluates two models, Qwen2.5-0.5B and Qwen2.5-3B \cite{qwen2024qwen25}. Fine-tuning uses parameter-efficient LoRA/QLoRA adaptation \cite{dettmers2023qlora} with rank 16, alpha 32, dropout 0.05, and the attention and feed-forward projection modules specified in the project configuration. The training setup uses 4-bit NF4 quantization, a per-device batch size of four with four gradient accumulation steps. It uses three epochs, learning rate $2\times10^{-4}$, maximum sequence length 1024, and gradient checkpointing. The experiments use sequential adapter initialization ($\texttt{previous\_adapter}$), such that for $g > 0$, model $M_g$ is initialized with the adapter produced by $M_{g-1}$.

Synthetic and prediction answer generation use greedy decoding. The maximum number of new tokens is set to 256, with a repetition penalty of 1.15, and a no-repeat 3-gram constraint. These settings are fixed across all generations and for both seeds 42 and 123. The same evaluation set and generation setup are used for both seeds.

\subsection{Evaluation metrics}
\paragraph {Biomedical entities} 
The \texttt{HFpf/\allowbreak{}pubmedbert-\allowbreak{}bc5cdr-\allowbreak{}ner}
model is used to process synthetic and reference answers. It is a PubMedBERT-based NER model fine-tuned on BC5CDR for disease and chemical entity recognition. Disease F1 and chemical F1 are computed by comparing normalized sets of extracted entities between synthetic and reference answers. Entity matches require the same entity type and identical normalized surface-form text. Disease and chemical entity retention are computed separately as the recall of reference entities that are present in answers. Under the exact normalized surface-form matching rule, entity retention is equal to reference-entity recall. F1 also includes precision to account for additional predicted entities.

Two entity-F1 summaries are used in the analysis. Table~\ref{tab:main_deltas}
reports the mean per-sample entity F1 over the evaluation set containing at least one reference entity, while Figure~\ref{fig:crossmodeltraj} reports micro-F1 computed from pooled entity counts across the evaluation set. The paired inferential analyses use the per-sample F1 values.

\paragraph{Context-supported rate}
For each predicted biomedical entity, the normalized surface-form is checked against the corresponding biomedical context. For each evaluation answer containing at least one biomedical entity, the context-supported rate is the fraction of predicted entities whose normalized surface-form occurs in the biomedical context. Generation-level values are reported as the mean across evaluable answers. Entities that do not appear in the context are reported as context-unmatched entities. They are not treated as definitive evidence of clinical hallucination.

\paragraph{Lexical and semantic similarity}
The evaluation includes BLEU-1, BLEU-4, ROUGE-L, METEOR, and chrF++ for lexical similarity. The evaluation also includes BERTScore F1 and embedding cosine similarity for semantic similarity. Embedding cosine similarity is computed using \texttt{sentence-transformers/all-mpnet-base-v2}.

\paragraph{Behavioral metrics}
For behavioral metrics, the study reports answer length, 3-gram repetition rate, and answer-only conditional perplexity. Answer-only conditional perplexity is computed over the answer tokens conditioned on a fixed evaluation prompt. Perplexity is used as a descriptive metric of generated text and is not considered a direct measure of biomedical answer quality.

\subsection{Statistical analysis}
For each training condition and seed, within-condition generation changes are computed on the paired per-sample evaluation table for $G_0$--$G_3$. The mean change, bootstrap 95\% confidence intervals, Cohen's $d_z$, and two-sided Wilcoxon signed-rank tests are computed. Selected inferential results are reported in the main text, while the complete statistical outputs are included in the released analysis artifacts.

The primary descriptive condition contrast is the difference-in-change value between Recursive and Human-Control:
\begin{equation}
\Delta_{RH}=\Delta_{R}-\Delta_{H}
\end{equation}
where $\Delta_R=R_{G_3}-R_{G_0}$ and $\Delta_H=H_{G_3}-H_{G_0}$. This is calculated separately for each seed. The results are then summarized using the mean across the two seeds. Because only two seeds are used for each model size, comparisons across conditions and model sizes are mainly descriptive rather than broader inferential comparisons.

\section{Results}

\begin{table*}[t]
\centering
\caption{Mean per-sample $G_3-G_0$ changes across Seeds 42 and 123. Negative values indicate a decrease in the metric. The final column within each model is the descriptive difference-in-change between the Recursive and Human-Control conditions.}
\label{tab:main_deltas}
\scriptsize
\begin{tabular}{lrrrrrr}
\toprule
& \multicolumn{3}{c}{Qwen2.5-0.5B} & \multicolumn{3}{c}{Qwen2.5-3B}\\
Metric & Recursive & Human-Control & $\Delta_{RH}$ & Recursive & Human-Control & $\Delta_{RH}$\\
\midrule
Disease F1 & -0.1008 & -0.0093 & -0.0915 & -0.1342 & +0.0103 & -0.1444\\
Chemical F1 & -0.0857 & +0.0340 & -0.1198 & -0.0977 & +0.0325 & -0.1302\\
Context-supported rate & -0.1649 & +0.0538 & -0.2187 & -0.2412 & +0.0855 & -0.3268\\
ROUGE-L & -0.0462 & -0.0062 & -0.0400 & -0.0555 & +0.0066 & -0.0620\\
BERTScore F1 & -0.0151 & -0.0120 & -0.0031 & -0.0275 & -0.0100 & -0.0175\\
Cosine similarity & -0.0631 & -0.0215 & -0.0416 & -0.0800 & -0.0137 & -0.0663\\
Answer length (words) & +5.79 & +4.21 & +1.58 & +18.19 & -4.63 & +22.82\\
3-gram repetition & -0.00020 & +0.00031 & -0.00052 & -0.00017 & +0.00020 & -0.00037\\
Perplexity & +9.8412 & +11.8536 & -2.0123 & +7.7118 & +3.2068 & +4.5050\\
\bottomrule
\end{tabular}
\end{table*}

\subsection{Overall recursive changes}
\begin{figure*}[t]
\centering
\includegraphics[width=0.96\textwidth]{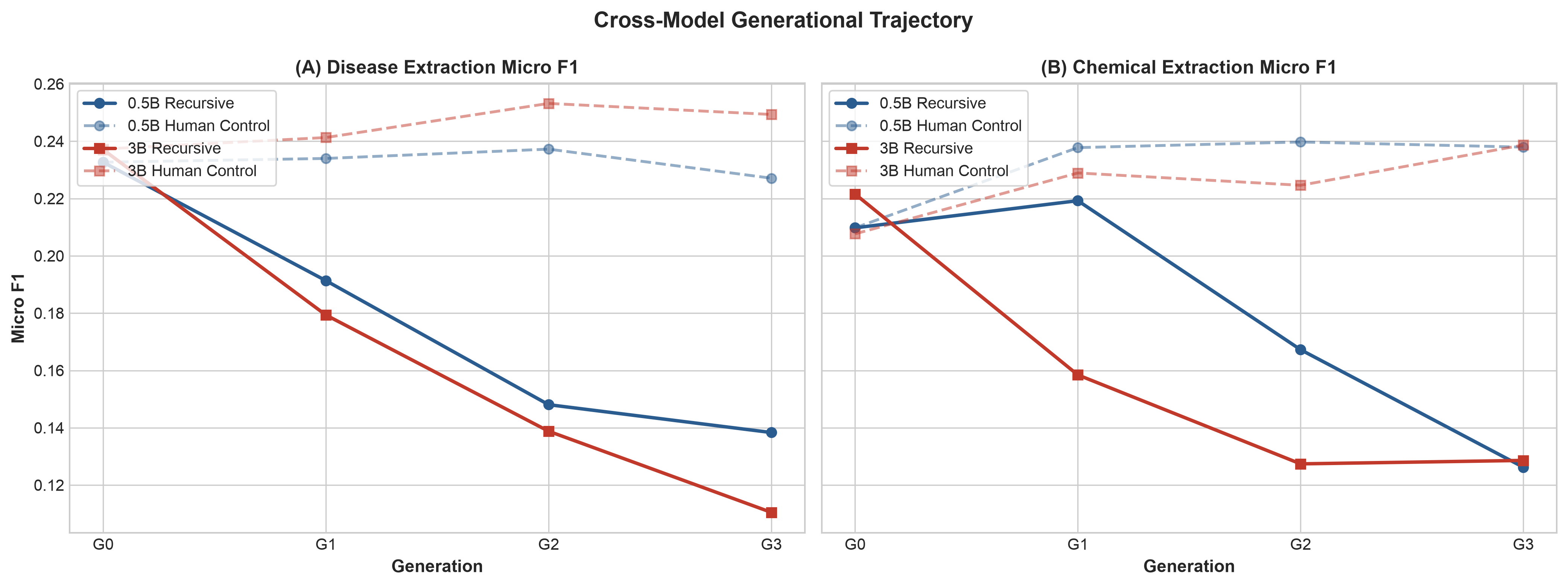}
\caption{Cross-model generational changes for disease and chemical entity micro-F1 across $G_0$--$G_3$. Lines show the mean across Seeds 42 and 123. Solid lines denote Recursive synthetic-data training, and dashed lines denote Human-Control; the two colors distinguish the 0.5B and 3B model configurations.}
\label{fig:crossmodeltraj}
\end{figure*}

The recursively trained models show larger declines compared to those of Human-Control models on the biomedical and reference-answer alignment metrics. Table~\ref{tab:main_deltas} reports the average $G_3-G_0$ change across seeds 42 and 123. The same pattern is observed in both model sizes for disease F1, chemical F1, context-supported rate, ROUGE-L, and cosine similarity. The changes in disease and chemical entity across $G_0$–$G_3$ for both model sizes and training conditions are shown in Figure~\ref{fig:crossmodeltraj}.

For Qwen2.5-0.5B, Recursive synthetic-data training reduces disease F1 by 0.1008 and context-supported rate by 0.1649. ROUGE-L and cosine similarity fall by 0.0462 and 0.0631, respectively. The corresponding changes in the Human-Control condition are smaller for disease F1, context-supported rate, ROUGE-L, and cosine similarity. The resulting difference-in-change values between the Recursive and Human-Control condition changes are -0.0915, -0.2187, -0.0400, and -0.0416, respectively.

The same pattern is larger for Qwen2.5-3B. While ROUGE-L decreases by 0.0555 and cosine similarity by 0.0800, Recursive disease F1 decreases by 0.1342 and context-supported rate by 0.2412. However, Human-Control disease F1 and ROUGE-L slightly increase over the same interval.

\begin{figure*}[t]
\centering
\includegraphics[width=0.96\textwidth]{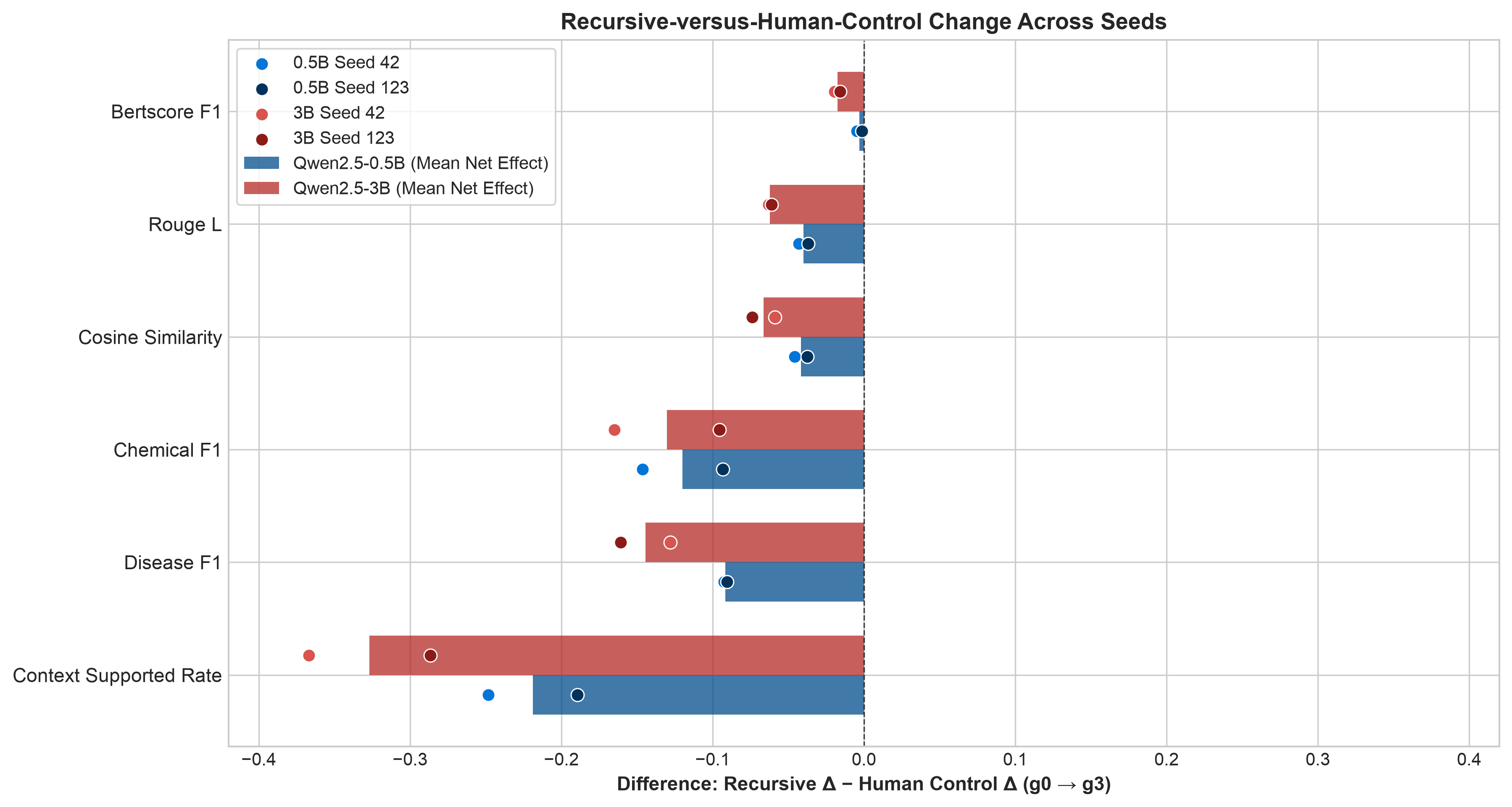}
\caption{Descriptive comparison of difference-in-change between the Recursive and Human-Control conditions in the core metrics from $G_0$ to $G_3$. Bars show the mean changes across Seeds 42 and 123, and the points show the two seed-specific changes. Negative values indicate a larger decline under Recursive synthetic-data training.}
\label{fig:conditioneffect}
\end{figure*}

\subsection{Seed consistency}
The direction of the difference-in-change between the Recursive and Human-Control condition is reproduced across both random seeds. For 0.5B, disease F1 difference-in-change values are -0.0925 (Seed 42) and -0.0904 (Seed 123), context-supported rate  values are -0.2481 and -0.1894, ROUGE-L values are -0.0431 and -0.0369. For 3B, corresponding disease F1 difference-in-change values are -0.1280 and -0.1609, context-supported rate difference-in-change values are -0.3670 and -0.2866, and ROUGE-L values are -0.0630 and -0.0611. This seed-level agreement is visible in Figure~\ref{fig:conditioneffect}, where the two seed-specific changes remain directionally consistent for the biomedical metrics within the tested configurations.

\subsection{Within-condition inferential results}
The paired per-example analyses provide statistically detectable changes in several metrics from $G_0$--$G_3$ within the Recursive condition. In the 3B Recursive condition, the disease F1 changes yielded small Wilcoxon $p$-values in both seeds ($p=0.00093$ and $p=0.00051$), while ROUGE-L, BERTScore F1, and cosine similarity also showed small $p$-values and large paired Cohen's $d_z$ values in both seeds. For example, the ROUGE-L Cohen's $d_z$ values are -0.887 and -0.862, and the corresponding BERTScore F1 values are -0.960 and -0.920. In the 0.5B Recursive condition, ROUGE-L, BERTScore F1, and cosine similarity likewise showed small $p$-values in both seeds, while disease and chemical F1 have wider uncertainty due to the smaller number of examples with evaluable entities.

These $p$-values should not be interpreted as direct tests of difference-in-change between the Recursive and Human-Control differences. The difference-in-change values between Recursive and Human-Control conditions are summarized by \(\Delta_{RH}\) above. Because only two seeds are used for each model size, comparisons across conditions and model sizes are mainly descriptive rather than broader inferential tests. The reported Wilcoxon \(p\)-values are exploratory within-condition tests and are not adjusted for multiple comparisons.

\subsection{Answer-length expansion without increased repetition}
A distinctive result is the divergence between answer length and reference-answer alignment. For the 3B model, recursive synthetic-data training increased the mean answer length from 42.18 words at $G_0$ to 60.37 words at $G_3$ (+18.19 words). In contrast, Human-Control reduced the mean answer length from 42.56 to 37.93 words (-4.63 words). Overall, the difference-in-change between the Recursive and Human-Control conditions was +22.82 words.

Under the fixed no-repeat 3-gram decoding constraint, the measured 3-gram repetition rate does not increase meaningfully in either model. Its difference-in-change between the Recursive and Human-Control conditions is also small (about -0.0005 for 0.5B and -0.0004 for 3B). These results suggest that, under the tested decoding configuration, the observed changes are more closely associated with content and context than with measured 3-gram repetition.

\subsection{Cross-Model difference-in-change between Recursive and Human-Control conditions}
Figure~\ref{fig:conditioneffect} summarizes the descriptive difference-in-change between the Recursive and Human-Control conditions. The values for difference-in-change were more negative for the 3B model than for the 0.5B model across disease F1 ($-0.1444$ vs.\ $-0.0915$), context-supported rate ($-0.3268$ vs.\ $-0.2187$), ROUGE-L ($-0.0620$ vs.\ $-0.0400$), BERTScore F1 ($-0.0175$ vs.\ $-0.0031$), and cosine similarity
($-0.0663$ vs.\ $-0.0416$). Answer length showed the largest behavioral
contrast, with a difference-in-change between the Recursive and Human-Control change of $+22.82$ words for the 3B model compared with $+1.58$ words for the 0.5B model.

The context-supported rate provides an additional domain-specific metric.
For the 0.5B model, the difference-in-change between the Recursive and Human-Control conditions was $-0.2187$. The seed-specific changes were $-0.2481$ and $-0.1894$ for
Seeds 42 and 123, respectively. For the 3B model, the corresponding
change was $-0.3268$. The seed-specific changes were $-0.3670$ and
$-0.2866$. Thus, both seeds showed lower context-supported rates under
Recursive synthetic-data training than under Human-Control at $G_3$.

The two seed-specific changes were directionally consistent across the reported metrics for both model sizes. These results provide a consistent descriptive comparison across the two model sizes within the tested configurations. The analysis therefore reports the observed cross-model contrasts without making a broader scaling claim.

\section{Discussion}
The results present a specific pattern of recursive degradation in biomedical QA. The feedback loop does not primarily produce shorter or obviously repetitive answers. Instead, answers become longer while becoming less similar to the original reference answers and less supported by the supplied biomedical context. This pattern is consistent with the intuition that a synthetic-data feedback loop can progressively alter model behavior relative to the original data distribution, but the present experiment shows the shift through domain-specific biomedical metrics rather than through a generic language-model benchmark.

The Human-Control condition is important for interpretation. Repeated fine-tuning itself can produce measurable changes. However, the scale and direction of the main changes are different when the training inputs remain the original human-authored data. The Human-Control therefore helps distinguish changes associated with using recursive synthetic-data feedback from changes associated with repeated fine-tuning alone. It does not eliminate all sources of generation-to-generation variation.

The larger-magnitude difference-in-change values are in the 3B configuration are also noteworthy. Larger model size is not demonstrated here to universally increase vulnerability. Rather, under the specific Qwen2.5 configurations tested, the 3B model shows a larger separation between Recursive and Human-Control conditions for several biomedical and semantic metrics, together with much stronger answer-length expansion. This model-size pattern should therefore be interpreted as a configuration-specific observation rather than as a general scaling relationship.

The findings demonstrate the value of using a variety of metrics for assessment. Lexical similarity captures a surface-level metric, while BERTScore and embedding cosine similarity capture broader semantic similarity. Entity F1 focuses on biomedically relevant terms, and context-supported rate metrics whether the biomedical entities in answers are supported by the input context. Relying on a single similarity metric alone might not be able to identify changes in the answers because each metric captures a different aspect of the answer.

These evaluation metrics should be viewed as complementary indicators rather than direct metrics of biomedical correctness. In particular, an entity that is not matched to the reference or context is not necessarily biomedically incorrect, and lexical or semantic differences do not necessarily imply factual degradation.

\section{Limitations}
The study tests with only two model sizes from the Qwen2.5 family and two random seeds. Hence, the obtained results can be used to confirm the directional consistency within the tested configurations, not universal scaling behavior.

Entity reference-answer mismatch, context-unmatched, and lexical and semantic similarity entities are indirect metrics. They do not necessarily mean that an answer contains a clinical hallucination or is biomedically incorrect without an expert analysis. Generation uses a no-repeat 3-gram decoding constraint, so the repetition results apply only to this decoding setting.

The training data for each generation is limited to 5,000 examples and is fine-tuned for three epochs. The reported changes may in turn reflect the dynamics of low-resource fine-tuning in addition to recursive synthetic-data loops, and results might differ under larger training data or more training epochs. 

The comparisons of the conditions and the model size depend upon just two seeds, which makes these findings primarily descriptive. The paired Wilcoxon and bootstrap analyses in the current results test within-condition generation changes. They should not be presented as a direct comparison of the difference-in-change between Recursive and Human-Control contrast.

\section{Conclusion}
This study shows that recursive synthetic-data training on synthetic biomedical QA answers produces a similar pattern of degradation in the preservation of biomedical entities, context-supported rate, and reference-answer alignment in four generations. Across seeds 42 and 123, the Recursive condition shows a consistent downward declines in the principal biomedical and reference-answer alignment metrics. Relative to Human-Control, the declines are evident in disease and chemical entity F1, context-supported rate, ROUGE-L, and cosine similarity.

The changes are larger in tested 3B for disease F1, context-supported rate, cosine similarity, and answer length. The degradation is not accompanied by a similar increase in simple 3-gram repetition. Instead, the results tend to show longer but less aligned answers across generations.

The results reflect the changes of recursive synthetic-data training in a biomedical QA setting rather than as proof of universal model collapse. Further work should evaluate additional human-anchored mixtures, additional model families and seeds, and direct expert assessment of biomedical factuality.

\section{Research Materials and Reproducibility}

The research code, experiment configurations, Jupyter notebooks, and plotting scripts can be found on 
\href{https://github.com/Anonymous-Chris/recursive-llm-degradation-research}{GitHub}, and the derived datasets, prediction results, and adapter weights for two models can be found on Hugging Face:

\noindent
\href{https://huggingface.co/chrislimbe/pubmedqa-recursive-llm-degradation-qwen2.5-0.5b-adapters}{Qwen2.5-0.5B Adapters},
\href{https://huggingface.co/chrislimbe/pubmedqa-recursive-llm-degradation-qwen2.5-3b-adapters}{Qwen2.5-3B Adapters},
\href{https://huggingface.co/datasets/chrislimbe/pubmedqa-recursive-llm-degradation-qwen2.5-0.5b}{Qwen2.5-0.5B Dataset},
\href{https://huggingface.co/datasets/chrislimbe/pubmedqa-recursive-llm-degradation-qwen2.5-3b}{Qwen2.5-3B Dataset}.

These releases contain derived research results and do not redistribute the original PubMedQA dataset in full. The released analysis results include per-sample evaluation outputs, generation-level summaries, statistical analyses, and plotting data used to produce the reported tables and figures.

\balance
% All entries in references.bib are the selected 20-paper bibliography.\n\nocite{*}
\bibliographystyle{IEEEtran}
\bibliography{references}

\end{document}